\documentclass[10pt,twocolumn,letterpaper]{article}

\usepackage{cvpr}
\usepackage{times}
\usepackage{epsfig}
\usepackage{graphicx}
\usepackage{amsmath}
\usepackage{amssymb}

\usepackage{tablefootnote}
\usepackage{multirow}
\usepackage{xcolor} 
\usepackage{verbatim}
\usepackage{url}
\usepackage{amssymb}
\usepackage{mathrsfs}

\usepackage{adjustbox}
\usepackage{diagbox}
\usepackage[pagebackref=true,breaklinks=true,letterpaper=true,colorlinks,bookmarks=false]{hyperref}

\cvprfinalcopy 

\def\cvprPaperID{5939} 

\ifcvprfinal\fi
\begin{document}

\title{Cascaded Refinement Network for Point Cloud Completion}

\author{Xiaogang Wang\quad\quad Marcelo H Ang Jr\quad\quad Gim Hee Lee\\
National University of Singapore\\
{\tt\small xiaogangw@u.nus.edu\quad \{mpeangh,gimhee.lee\}@nus.edu.sg}}

\maketitle

\begin{abstract}
Point clouds are often sparse and incomplete. Existing shape completion methods are incapable of generating details of objects or learning the complex point distributions. 
To this end, we propose a cascaded refinement network together with a coarse-to-fine strategy to synthesize the detailed object shapes. Considering the local details of partial input with the global shape information together, we can preserve the existing details in the incomplete point set and generate the missing parts with high fidelity. We also design a patch discriminator that guarantees every local area has the same pattern with the ground truth to learn the complicated point distribution. Quantitative and qualitative experiments on different datasets show that our method achieves superior results compared to existing state-of-the-art approaches on the 3D point cloud completion task. Our source code is available at \url{https://github.com/xiaogangw/cascaded-point-completion.git}.

\end{abstract}

\section{Introduction}

Despite the significant progress on image generation and translation~\cite{wang2018pix2pixHD,isola2017image}, synthesizing and generating 3D point clouds remains as a very challenging task due to the sparseness,  incompleteness and irregularity of the points. 
More specifically, the inabilities of learning accurate point features and various point distributions make it difficult to obtain a complete and dense object shape.
In this work, we focus on the point cloud completion~\cite{yuan2018pcn,topnet2019} task, which completes missing parts of the occluded object. 3D shape completion has wide applications such as robotic navigation~\cite{engel2014lsd,mur2015orb}, scene understanding~\cite{hou2019sis,dai2018scancomplete} and augmented reality~\cite{boud1999virtual,webster1996augmented}. 

\begin{figure}
  \includegraphics[width=\linewidth]{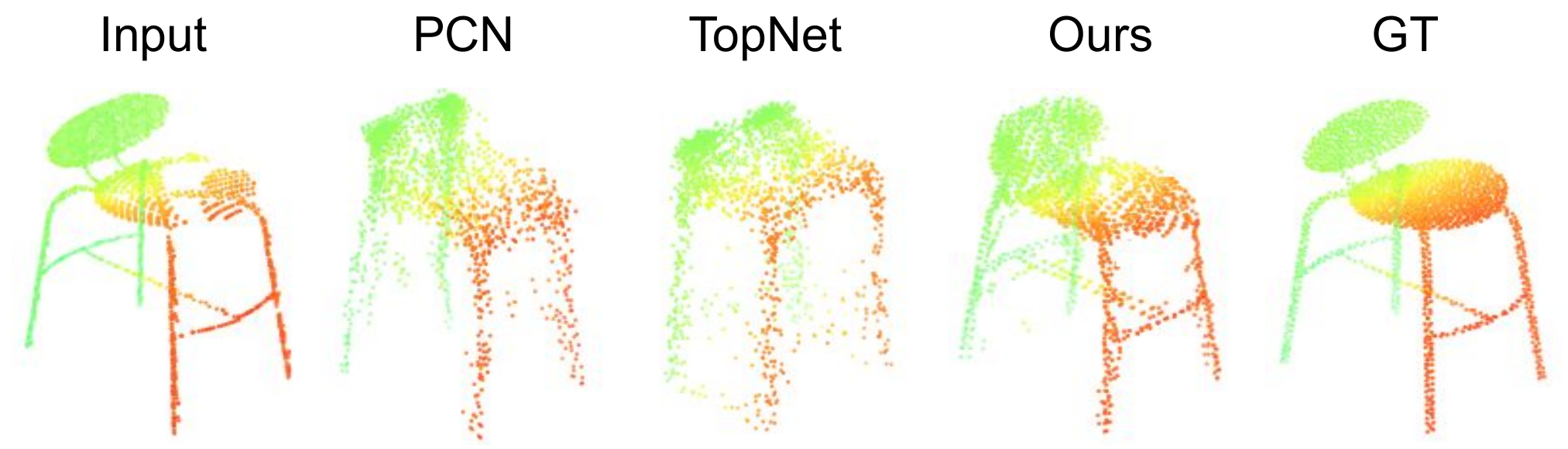}
  \caption{Our method can generate complete point clouds with finer details compared to existing state-of-the-art methods.}
  \label{task_picture}
  \vspace{-0.6cm}
\end{figure}

Existing methods~\cite{yuan2018pcn,topnet2019,brock2016generative,dai2017shape,litany2018deformable,sinha2017surfnet} have shown promising results on shape completion for different inputs: distance fields, meshes, voxel grids and point clouds. Voxel representations are a direct generalization of pixels to the 3D case. However, generating 3D shape with voxel format suffers from memory inefficiency, hence it is difficult to obtain high-resolution results. 
Although data-driven methods on mesh representations~\cite{wang2018pixel2mesh,groueix2018atlasnet,wang20193dn} are able to generate complicated surfaces, they are limited to the fixed vertex connection patterns. As a result, it is difficult to change the topology during the training process. 
In contrast, it is easy to add new points for point clouds and several studies have shown promising results.
The pioneer work~\cite{yuan2018pcn} proposes an encoder-decoder based pipeline on both the synthetic dataset ShapeNet~\cite{chang2015shapenet} and the real scene dataset KITTI~\cite{geiger2013vision}. A following work TopNet~\cite{topnet2019} proposes a hierarchical rooted tree structure decoder to generate object shapes. Even though they have achieved impressive performances on shape completion, they are both unable to generate the detailed geometric structure of 3D objects and produce unsatisfactory coarse object outputs. 
Several approaches~\cite{mescheder2019occupancy,park2019deepsdf,michalkiewicz2019deep} propose to learn 3D structures in a function space, and they achieve impressive results for various input formats. However, these methods require post-processing to refine the outputs.

We propose to synthesize the dense and complete objects shapes in a cascaded refinement manner, and jointly optimize the reconstruction loss and an adversarial loss end-to-end.
Our framework is designed to keep the object details in the partial inputs, and to produce realistic reconstructions of the missing parts. 
Figure~\ref{task_picture} shows an example between our method and existing approaches~\cite{yuan2018pcn,topnet2019}. Although the legs of a chair are clearly present in the input, existing works are incapable of keeping this structural details in the outputs.
On the contrary, our approach successfully captures this fine-grained details.
To this end, we make a skip connection between the incomplete points and coarse outputs. However, simple concatenation between inputs and our coarse outputs give rise to unevenly distributed points. Consequently, we design an iterative refinement decoder together with a feature contraction and expansion unit to refine the point positions.
We adopt an adversarial loss that penalizes inaccurate points from the ground truth to learn the complex point distributions and further improve the performance. 
Instead of classifying the whole object by predicting a single confidence value like conventional generative adversarial networks (GANs)~\cite{li2019pu,wu2016learning}, we design a patch-based discriminator to explicitly force every local patch of  generated point clouds to have the same pattern with real complete point clouds inspired by ~\cite{isola2017image,wu2019point}.
We show state-of-the-art quantitative and qualitative results on different datasets by various experiments. 
Our key contributions are as follows:
\begin{itemize}
    \item We propose a novel point cloud completion network which is able to preserve object details from partial points and generate missing parts with fine details at the same time;
    \item Our cascaded refinement strategy together with the coarse-to-fine pipeline can refine the points positions locally and globally;
    \item Experiments on different datasets show that our framework achieves superior results to existing methods on the 3D point cloud completion task.
\end{itemize}

\begin{figure*}
\centering
  \includegraphics[width=1\linewidth]{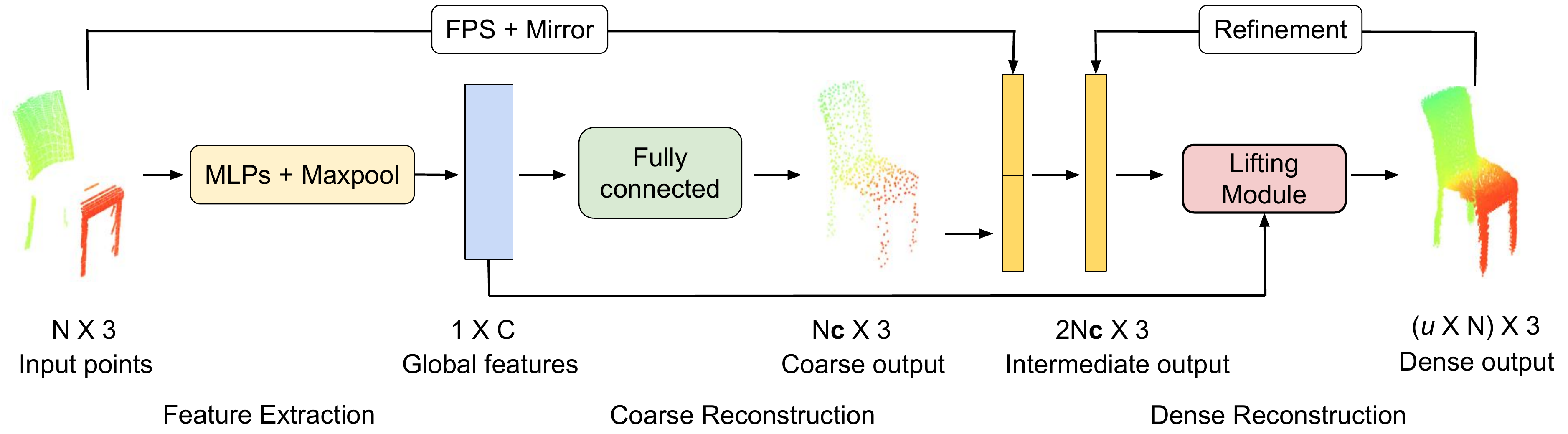}
  \caption{An illustration of our generator network. The generator includes three sub-networks: feature extraction, coarse reconstruction and dense reconstruction. The feature extractor consists of two MLPs and max-pooling layers. The coarse reconstruction comprises several fully-connected layers. The dense reconstruction is a cascaded refinement sub-network with a lifting module in each step. We generate dense and complete point clouds given partial and sparse inputs. $\mu$ is the upsampling factor.}
    \vspace{-0.2cm}
  \label{whole_network}
\end{figure*}

\section{Related work}

In this section, we review existing works on 
point generation, upsampling and shape completion that are related to our task.

\vspace{-0.3cm}
\paragraph{3D Generation.}
The pioneering work PointNet~\cite{qi2017pointnet} proposed a method on point cloud analysis and inspired a large amount of works on point cloud generation.
Early works~\cite{achlioptas2017learning,valsesia2018learning,shu20193d} have proposed generative models by using GAN or variational auto-encoder (VAE) on 3D generation. 
Achlioptas et al.~\cite{achlioptas2017learning} proposed r-GAN for 3D point clouds generation, in which both generator and discriminator are fully connected layers. 
Valsesia et al.~\cite{valsesia2018learning} proposed a graph neural network to synthesize object shapes. They calculated the adjacency matrix by the feature vectors from each vertex in each graph convolution layer. Despite their superior results, the calculation of the adjacency matrix requires quadratic computation complexity and consumes a lot of memory.
The above methods successfully synthesize object shapes from noise. However, simple GANs or VAE can only generate small scale (1024 or 2048) point sets due to the complex point distribution and the notoriously difficult training of GANs. 
Although improved methods~\cite{yang2019pointflow,shu20193d,yang2018foldingnet} show superior performance on synthesizing 3D objects, they are limited to synthesizing the general shapes of objects and are not suitable for shape completion. 

\vspace{-0.3cm}
\paragraph{3D Upsampling.}
Similar to point cloud completion, several works~\cite{yu2018pu,yu2018ec,Yifan_2019_CVPR,li2019pu,wu2019point} aim at generating dense and uniform point clouds given sparse and non-uniform point sets. 
PU-Net~\cite{yu2018pu} adopted the PointNet++~\cite{qi2017pointnet++} as a backbone to extract point features and expand feature dimensions by a series of convolutions. Following PU-Net, EC-Net~\cite{yu2018ec} generated sharp edges by penalizing the distance between points and edge labels. While they show exciting results, both methods are limited to upsampling point clouds by a small ratio (e.g. 4$\times$). To alleviate this problem, Yifan et al.~\cite{Yifan_2019_CVPR} introduced a hierarchical point feature extraction and multi-stage generation network and achieved 16$\times$ upsampling, but the training process consumes more computation memory.
More importantly, they are all limited to upsampling the sparse points and are not applicable for completion tasks.

\vspace{-0.3cm}
\paragraph{3D Completion.}
3D shape completion plays an important role in robotics and perception, and has obtained significant development in recent years. Existing methods have shown impressive performance on various formats: voxel grids, meshes and point clouds.
Inspired by 2D CNN operations, earlier works~\cite{dai2017shape,han2017high,stutz2018learning,le2018pointgrid} focus on the voxel and distance fields formats generation with 3D convolution. Several approaches~\cite{dai2017shape,stutz2018learning} have proposed a 3D encoder-decoder based network for shape completion and shown promising performance. However, voxel-based methods consume a large amount of memory and are unable to generate high-resolution outputs. To increase the resolution, several works ~\cite{wang2017cnn,wang2018adaptive} have proposed to use the octree structure to gradually voxelize specific areas. However, due to the quantization effect of the voxelization operation, recent works gradually discard the voxel format and focus on the mesh reconstruction. 
Existing mesh representations~\cite{groueix2018atlasnet,wang2018pixel2mesh} are based on deforming a template mesh to a target mesh and hence not flexible to any typologies. 
In comparison to voxels and meshes, point clouds are easy to add new points during the training procedure.
Yuan et~al.~\cite{yuan2018pcn} proposed the pioneering work PCN on point cloud completion, which was a simple encoder-decoder network to reconstruct dense and complete point set from an incomplete point cloud. They adopted the folding mechanism~\cite{yang2018foldingnet} to generate high resolution outputs (16,384).
TopNet~\cite{topnet2019} proposed a hierarchical tree-structure network to generate point cloud without assuming any specific topology for the input point set. 
However, both PCN and TopNet are unable to synthesize the fine-grained details of 3D objects.

\section{Our Method}
\subsection{Overview}
Our objective is to produce complete and high-resolution 3D objects from corrupted and low resolution point clouds. Specifically, given the sparse incomplete point sets $\text{P}=\{p_i\}^N_{i=1}$ of $N$ points, we aim to generate a dense and complete point set $\text{Q}=\{q_i\}^{u \times N}_{i=1}$ of $u  \times N$  points, where $u$ is the upsampling scalar. We expect our method to fulfill three requirements: (1) preserve the fine details of the input point cloud $\text{P}$, (2) inpaint the missing parts with detailed geometric structures, and (3) generate evenly distributed points on object surfaces.

Our point cloud completion architecture is shown in Figure~\ref{whole_network}. 
Traditional GANs \cite{goodfellow2014generative,achlioptas2017learning,valsesia2018learning} map a noise distribution $z$ to the data space, we extend the general GAN framework by modelling the generator $G$ (Section \ref{sect:generator}) as a feature extraction encoder and a conditional coarse-to-fine decoder.
The discriminator $D$ (Section \ref{sect:discriminator}) aims to distinguish between the generated fake output and the ground truth.

\subsection{Generator}\label{sect:generator}
Our generator $G$ consists of three components: (1) feature extraction $h$, (2) coarse reconstruction $g_1$ and (3) dense reconstruction $g_2$.

\vspace{2.5mm}
\noindent\textbf{Feature Extraction.}
Same with PCN~\cite{yuan2018pcn}, we use two stacked PointNet feature extraction architecture with max-pooling operation to extract the global point features $f$.
Specifically, the feature extractor $h$ can be modelled by the composition of two functions expressed as:
\begin{align}
    f= h(\text{P} \mid \text{w}_{h}), \quad
    h=h_1 \circ h_2, 
\end{align}
where $\text{w}_{h}$ denotes the parameters of $h_1$ and $h_2$, $h_1$ and $h_2$ represent the two extraction sub-networks, respectively. 

\vspace{2.5mm}
\noindent\textbf{Coarse Reconstruction.} $g_1$ consists of several fully-connected layers, which maps the latent embedding $f$ to the coarse point cloud. We denote the size of $\text{P}_{\text{coarse}}$ as $N_c\times 3$. From Figure~\ref{whole_network}, we can observe that the coarse output roughly capture the complete object shape but loses fine details, which we aim to recover in the second stage. 

\vspace{2.5mm}
\noindent\textbf{Dense Reconstruction.}
Our second stage $g_2$ is a conditional iterative refinement sub-network. 
The synthesis begins at generating low resolution points (2048$\times$3), and points with higher resolutions are then progressively refined. 
Following TopNet~\cite{topnet2019}, our outputs have four resolutions: $N=\{2048, 4096, 8192, 16,384\}$, for which the numbers of iterations are 1, 2, 3 and 4, respectively. 
Parameters are shared among each iteration.

Existing methods~\cite{yuan2018pcn,topnet2019,shu20193d} exploit either folding based operations or tree structure to generate dense and complete objects. 
Although they have achieved impressive qualitative results, the fine details of the objects are often lost. 
As can be seen in Figure~\ref{task_picture}, both PCN~\cite{yuan2018pcn} and TopNet~\cite{topnet2019} fail to generate the details of 3D objects (e.g. the legs of the chair).
The reason is that the latent embedding $f$ is obtained by the last max-pooling layer of the encoder, and it only represents the rough global shape, hence it is difficult to recover the detailed object structures. 
We propose to preserve the object shape details in the partial inputs and exploit the global shape information from $\text{P}_{\text{coarse}}$ at the same time.
Inspired by the skip-connection from U-Net~\cite{ronneberger2015u}, we concatenate the partial inputs with the global shape $\text{P}_{\text{coarse}}$ to synthesize the dense points.
However, direct concatenation resulted in a poor visual quality because of the serious uneven distributed points. 
To alleviate this problem, we propose to dynamically subsample $N_c\times 3$ points from the partial inputs $\text{P}$ before concatenating with the coarse output $\text{P}_{\text{coarse}}$. 
We denote the combined point sets as $\text{P}_S$ with the size of $2N_c\times 3$, which are fed into the lifting module (Section \ref{sect:Lifting_Module}) to obtain a higher resolution points $\text{P}_{\text{i}}$. We use the efficient farthest point sampling (FPS) algorithm~\cite{qi2017pointnet++} to subsample points.
We also design a feature contraction-expansion unit (Section \ref{sect:Lifting_Module}) to refine the point positions gradually.
We progressively refine the point positions and upsample the point size by a factor of two by the lifting module.
For the subsequent iterates, the input for the lifting module is the intermediate output $\text{P}_{\text{i}}$ from last step.

\vspace{-0.0cm}
\subsection{Lifting Module}\label{sect:Lifting_Module}
We design a lifting module to upsample the point size by a factor of two, and concurrently refine the point positions by the feature contraction and expansion unit.
To upsample the point set $\text{P}_S$, we first tile the points $\text{P}_S$ two times to obtain a new point set $\text{P}_S^{\prime}$. 
Then we sample a unique 2D grid vector and append it after each point coordinates to increase the variations among the duplicated points~\cite{yang2018foldingnet}.
We also utilize the mean shape prior $f_m$ (Section \ref{sect:Shape_Priors}) in our iterative refinement~\cite{kanazawa2018end} to alleviate the domain gap of point features between the incomplete and complete point clouds.
We concatenate the point $\text{P}_S^{\prime}$, mean shape vectors $f_m$, global feature $f$ and the sampled 2D grids to obtain a new feature $f_s$.
We aim to predict per-vertex displacements $\{d_x,d_y,d_z\}$  for each point $\text{P}_S^{\prime}$ given the point feature $f_s$. 

\begin{figure}
\centering
  \includegraphics[width=1\linewidth]{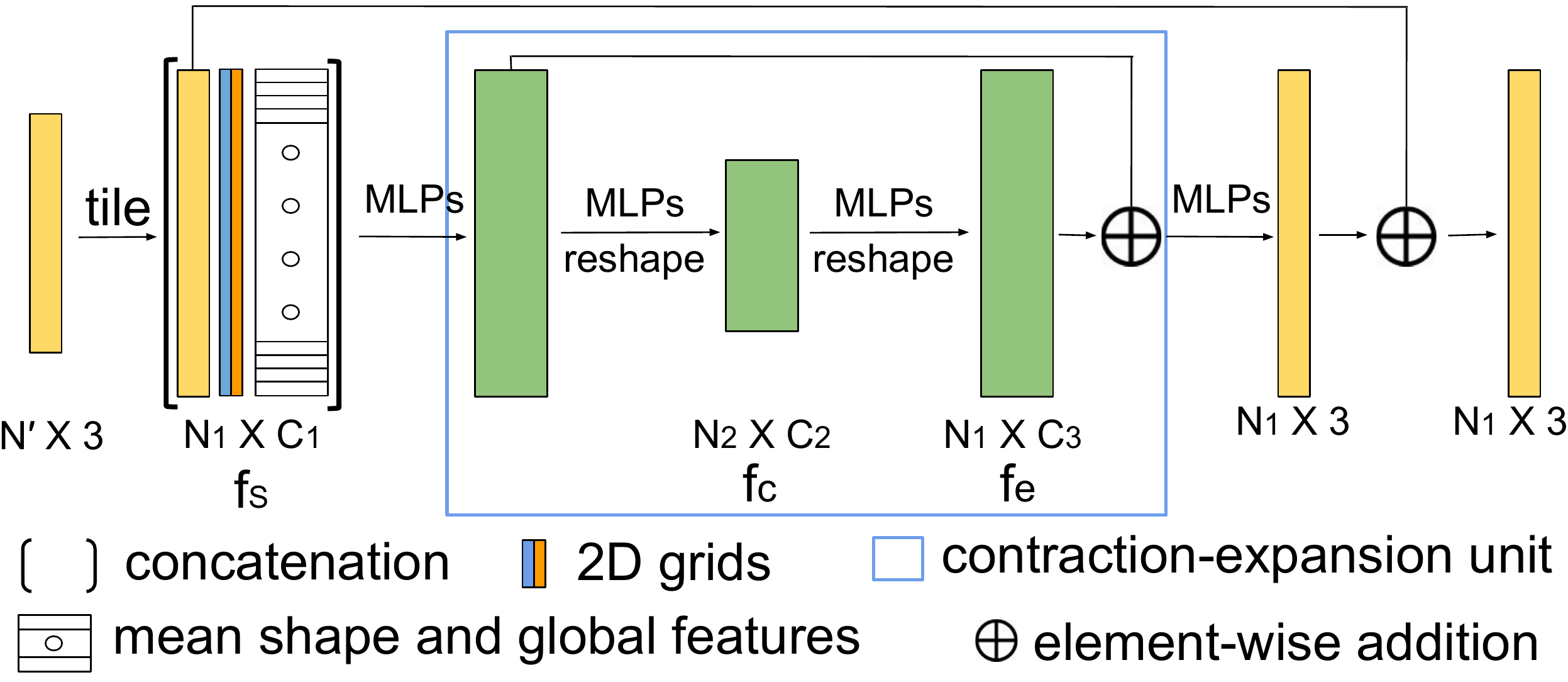}
  \caption{The architecture of the lifting module. The input is $N^{\prime}\times 3$, and we upsample it by a factor of 2 to obtain the output of $N_1\times 3$. The feature contraction and expansion unit predicts residual point features instead of final results.}
  \label{lifting_unit}
  \vspace{-0.3cm}
\end{figure}

\vspace{3mm}
\noindent\textbf{Feature Contraction-expansion Unit.}
Inspired by the hourglass network~\cite{newell2016stacked}, we consolidate the local and global information by a bottom-up and top-down fashion to refine points positions and make them evenly distributed on object surfaces.
However, it is not straightforward to subsample and upsample features between different scales for point clouds. Although some operations are introduced in PointNet++~\cite{qi2017pointnet++} and graph convolution~\cite{dgcnn}, they consume a large amount of memory and computation time, especially for high-resolution points. 
Consequently, we use shared multilayer perceptrons (MLPs)~\cite{yu2018pu} to make feature contraction and expansion.
Specifically, we assume the dimension of $f_s$ to be $N_1\times C_1$, and sizes of outputs features $f_c$ and $f_e$ are $N_2\times C_2$ and $N_1\times C_3$, respectively. The two operations are represented as $f_c=\mathcal{RS}(\mathcal C_{C}(f_S))$ and $f_e=\mathcal{RS}(\mathcal C_{E}(f_c))$,
where $\mathcal{RS}(.)$ is a reshaping operation. $\mathcal C_{C}(\cdot)$ and $\mathcal C_{E}(\cdot)$ are MLPs for contraction and expansion, respectively.
Our lifting module predicts point feature residuals rather than the final output since deep neural networks are better at predicting residuals~\cite{wang2018pixel2mesh}. 
Our lifting module is shown in Figure~\ref{lifting_unit}.

Overall, in one-step refinement, the output point set $\text{P}_{\text{i}}$ is represented as:
\begin{align}
    \text{P}_i&=F(\text{P}_S^{\prime})+\text{P}_S^{\prime},
\end{align}
where $F(.)$ predicts per-vertex offsets by the lifting module for the input point $\text{P}_S^{\prime}$.

\begin{figure}
\centering
  \includegraphics[width=1\linewidth]{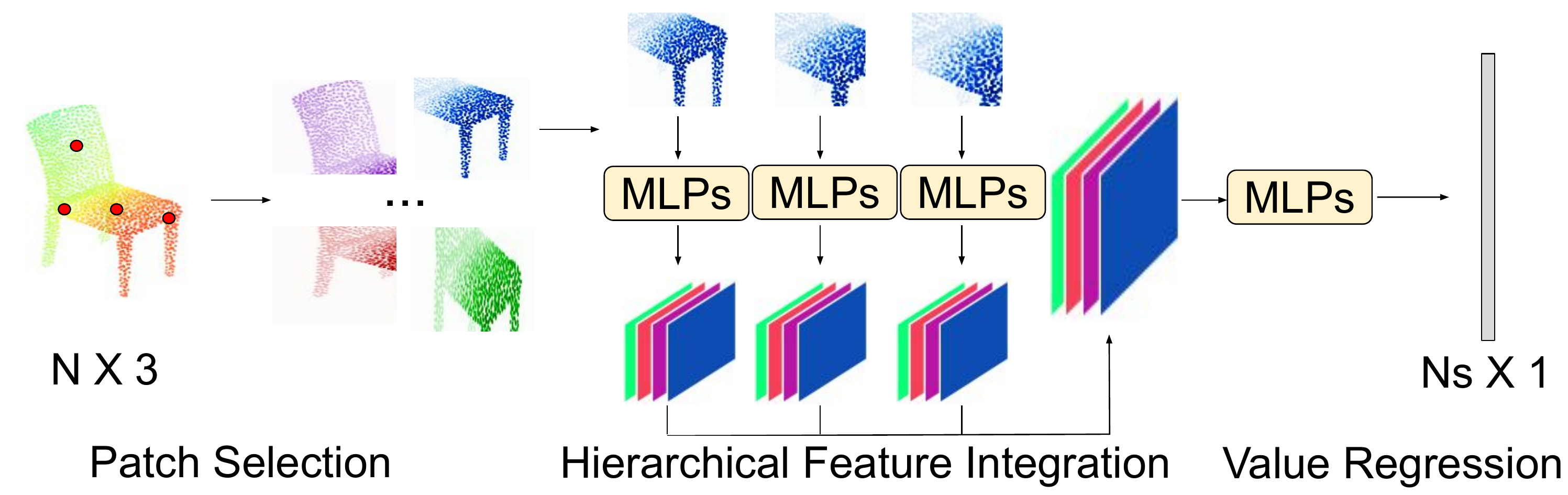}
  \caption{The discriminator architecture sub-network. It includes the patch selection, hierarchical feature integration and confidence value regression.}
    \vspace{-0.3cm}
  \label{discriminator}
\end{figure}


\subsection{Shape Priors}\label{sect:Shape_Priors}
For each object class, we take mean values of latent embeddings from all the instances within that category as our mean shape vectors. 
The calculation is represented as:
\begin{align}
    f_{m}^i=\frac{1}{N^i}\sum_{i=1}^{i=N^i} f_{i},
\end{align}
where $N^i$ is the total number of objects from category $i$. 
The latent embeddings are obtained by a pre-trained PointNet auto-encoder\footnote{https://github.com/charlesq34/pointnet-autoencoder} on eight object categories following~\cite{kanazawa2018end}.

Following 3DN~\cite{wang20193dn}, we mirror the partial input with respect to the $xy$-plane as we assume the reflection symmetry plane ($xy$-plane) of objects to be known since many man-made models show global reflection symmetry.
Therefore we mirror the subsampled input points to obtain the point set $\text{P}_C$ with the size of $N_c\times 3$.
Note that not all training objects are symmetric, and 40 of 1200 testing data are asymmetric. Our mirror operation can be seen as an initialization for the missing points and reasonable point positions are generated by the whole optimization.

\subsection{Discriminator}\label{sect:discriminator}
To generate various realistic dense and complete point clouds, we adopt the adversarial training and jointly optimize the reconstruction loss and the adversarial loss end-to-end. 
Instead of only considering the global shape by regressing one single confidence value as conventional GANs~\cite{li2019pu}, we design a patch discriminator to further guarantee that every local area is realistic.
We employ the hierarchical point set feature learning in PointNet++~\cite{qi2017pointnet++} with different radii to consider multi-scale local patches. Specifically, we first uniformly sample $N_s$ point seeds by FPS, and then set three radii $\{0.1, 0.2, 0.4\}$ around the seeds to extract a set of local patch. 
Finally, we obtain $N_s$ scores from the discriminator instead of calculating one single value for binary classification.
Our discriminator consists of patch selection, hierarchical feature integration and value regression. 
The discriminator sub-network is shown in Figure~\ref{discriminator}.

\subsection{Optimization}
Our training loss comprises two components, a reconstruction loss to encourage the completed point cloud to be the same as the ground truth, and an adversarial loss to penalize the unrealistic outputs.

\vspace{2.5mm}
\noindent\textbf{Reconstruction Loss.}
We adopt the Chamfer Distance (CD) \cite{fan2017point} as our reconstruction loss, i.e., 
\begin{align}
    \begin{split}
    \text{CD}(\text{X},\text{Y})&=\mathcal{L}_{\text{X},\text{Y}}+ \mathcal{L}_{\text{Y}, \text{X}},~ \text{where}\\
    \mathcal{L}_{\text{X},\text{Y}}&=\frac{1}{|\text{X}|}\sum_{x\in \text{X}}  \min\limits_{y\in \text{Y}}||x-y||_2,~\text{and}\\
    \mathcal{L}_{\text{Y},\text{X}}&=\frac{1}{|\text{Y}|}\sum_{y\in \text{Y}}  \min\limits_{x\in \text{X}}||x-y||_2,\\
    \end{split}
\end{align}
which calculates the average closest point distance between two point clouds $\text{X}$ and $\text{Y}$.
There are two variants for CD which we denote as CD-P and CD-T. Specifically,
CD-P takes the square root operation and is divided by 2. 
We show different results with these two variants, and we adopt $\text{CD}\text{-}\text{P}$ in all our experiments during training.
Hence, our reconstruction loss can be expressed as:
\begin{align}
    \mathcal{L}_{\text{rec}}=\text{CD}(\text{P}_{\text{coarse}},\text{Q})+\lambda_\text{f} \text{CD}(\text{P}_{\text{fine}},\text{Q}),
\end{align}
where $\text{P}_{\text{coarse}}$ and $\text{P}_{\text{fine}}$ correspond to the coarse output and fine output, respectively, and $\lambda_\text{f}$ is the weight for the reconstruction loss of $\text{P}_{\text{fine}}$.

\vspace{2.5mm}
\noindent\textbf{Adversarial Loss.}
We adopt the stable and efficient objective function of LS-GAN~\cite{mao2017least} for our adversarial losses. Specifically, the adversarial losses for the generator and discriminator are:
\begin{align}
    \mathcal{L}_{\text{GAN}}(G) = \frac{1}{2}[D(\tilde{x})-1]^2,
\end{align}
\begin{align}
    \mathcal{L}_{\text{GAN}}(D) = \frac{1}{2}[D(\tilde{x})^2 +(D(x)-1)^2 ],
\end{align}
where $\tilde{x}$ and $x$ are the generated fake result and the target ground truth, respectively. 

\vspace{2.5mm}
\noindent\textbf{Overall Loss.}
Our overall loss function is the weighted sum of the reconstruction loss and the adversarial losses:
\begin{align}
    \mathcal{L}=\lambda \mathcal{L}_{\text{GAN}} + \beta \mathcal{L}_{\text{rec}},
\end{align}
where $\lambda$ and $\beta$ are the weights for GAN loss and the reconstruction loss, respectively.
During training, $G$ and $D$ are optimized alternatively.

\begin{figure*}
	\centering
	\includegraphics[width=1\linewidth]{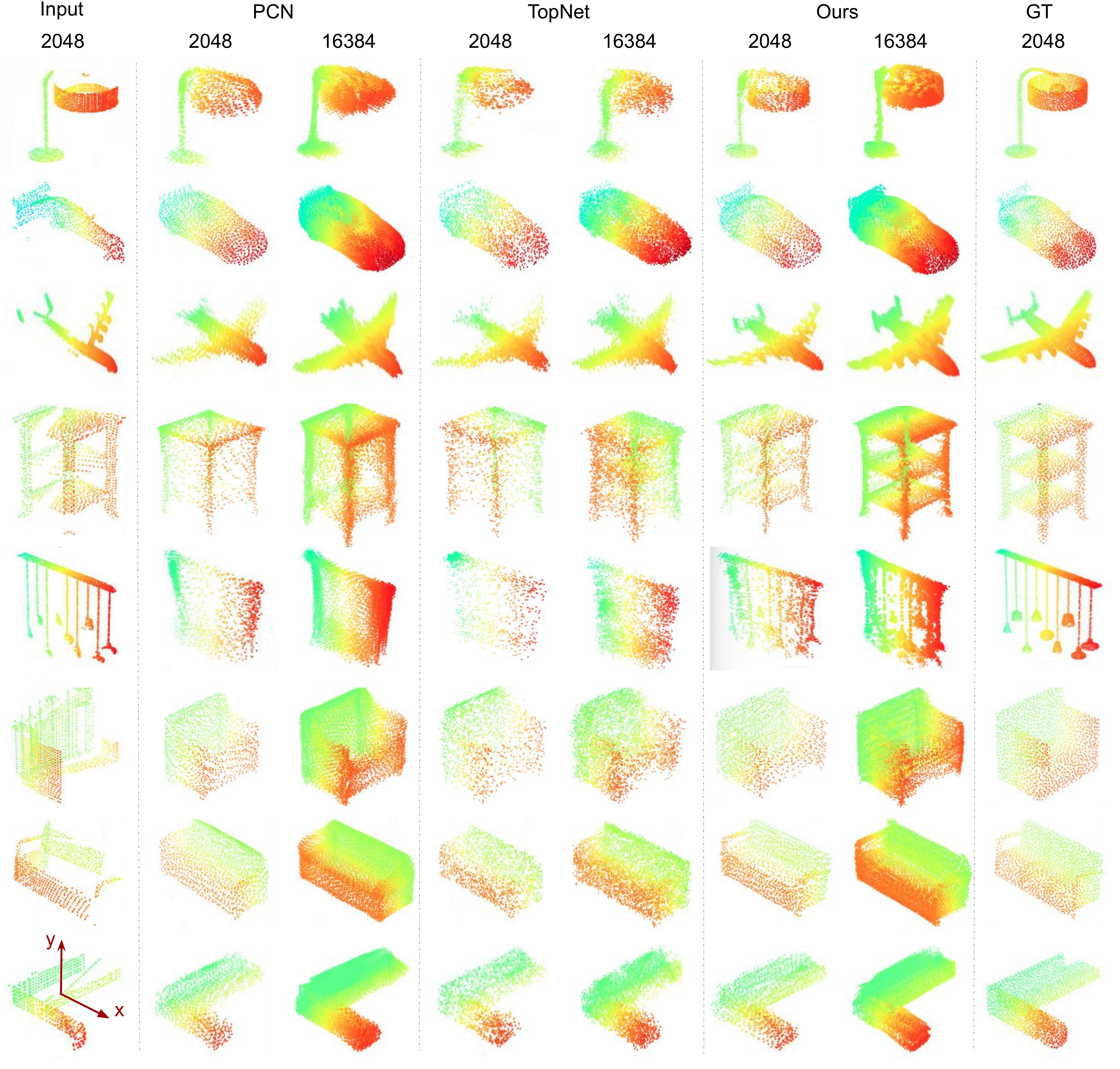}
	\caption{Qualitative comparison on our created ShapeNet dataset. The resolution for both partial and ground truth are 2048. We show the generated results of size 2048 and 16,384 from different methods.}
	\label{qualitative_results_2048}
	\vspace{-0.3cm}
\end{figure*}

\section{Experiments}

\subsection{Evaluation Metrics}
We compare our method with several existing methods 3D-EPN~\cite{dai2017shape}, PCN~\cite{yuan2018pcn} and TopNet~\cite{topnet2019}.
We use two evaluation metrics to evaluate results quantitatively. The first metric is the Chamfer Distance (CD) following~\cite{yuan2018pcn,topnet2019}. More specifically, we use CD-P for experiments in Section~\ref{exp_pcn} and use CD-T in the remaining experiments for fair comparison.
The other metric is Fr$\acute{\text{e}}$chet Point Cloud Distance (FPD) adopted from~\cite{shu20193d}.
FPD calculates the 2-Wasserstein distance between the real and fake Gaussian measures in the feature spaces of the point sets:
\begin{equation}
    \begin{split}
    \text{FPD}(\text{X},\text{Y})=\|\text{m}_{\text{X}}-\text{m}_\text{Y}\|_2^2 +
    \text{Tr}(\Sigma_{\text{X}}+\Sigma_{\text{Y}} -2(\Sigma_{\text{X}}\Sigma_\text{Y})^{\frac{1}{2}}),
    \end{split}
\end{equation}
where m and $\Sigma$ represent the mean vector and covariance matrix of the points, respectively. Tr($A$) is the sum of the diagonal elements from matrix $A$. More evaluation details are shown in supplementary material.

\subsection{Datasets}
For a fair comparison, we evaluate our method on the datasets of PCN~\cite{yuan2018pcn} and TopNet~\cite{topnet2019}. 
Partial inputs are obtained by back-projecting 2.5D depth images into 3D. 30,974 objects from eight categories are selected: airplane, cabinet, car, chair, lamp, sofa, table and vessel. 
We also create our smaller training dataset to measure the generalization ability on fewer training data. 
We only render the partial scans with one random virtual view instead of eight random views like PCN, hence the number of our training data is one eighth of PCN, but we keep the testing data the same with PCN. The resolutions of the partial and complete point clouds are 2048 in our created dataset following TopNet~\cite{topnet2019}.
We use our testing data for evaluation when training on the dataset of TopNet.

\subsection{Implementation Details}
All our models are trained using the Adam~\cite{kingma2014adam} optimizer.
We adopt the two time-scale update rule (TTUR)~\cite{heusel2017gans} and set learning rates for the generator and discriminator as 0.0001 and 0.00005, respectively. 
The learning rates are decayed by 0.7 after around every 40 epochs, and clipped by $10^{-6}$.
$\lambda$ and $\beta$ are set to 1 and 200, respectively.  $\lambda_\text{f}$ increases from 0.01 to 1 within the first 50,000 iterations. $N_s$ in discriminator is 256.
The size $N_c$ of coarse output is 512.
We train one single network for all eight categories of data. 

\subsection{Point Completion on the Dataset of PCN}\label{exp_pcn}
\begin{figure}
\centering
  \includegraphics[width=1\linewidth]{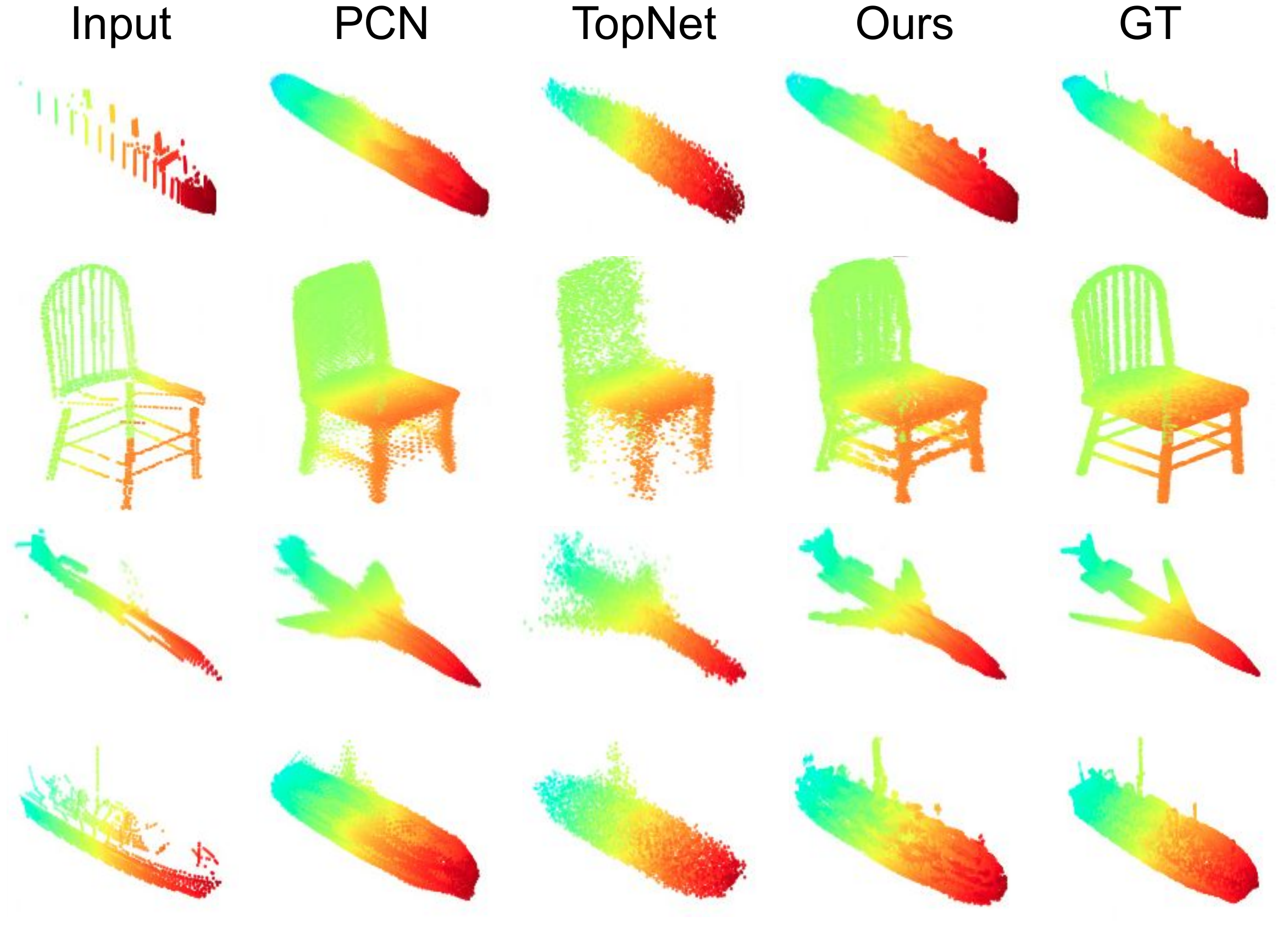}
  \caption{Qualitative comparison on the dataset of PCN. Point resolutions for the output and ground truth are 16,384.}
  \label{qualitative_results}
\end{figure}

Quantitative and qualitative results are shown in Table~\ref{quantative_seen} and Figure~\ref{qualitative_results}. 
Point resolutions for the output and the ground truth are 16,384.
The quantitative results in Table~\ref{quantative_seen} show that we obtain the best performance on all categories of objects compared to other methods.
We obtain $11.74\%$ relative improvement on the average value compared to the second best method PCN.  The results indicate that we achieve better performance with more accurate global shape and finer local structures.
From Figure~\ref{qualitative_results} we can observe that PCN and TopNet fail to recover the fine details such as legs of a chair and aircraft tails, while our method successfully generates such structures.

\begin{table*}[!htbp]
\centering
\resizebox{0.95\textwidth}{!}{
\small
\begin{tabular}{l|c| c |c  |c  | c|  c|  c |c|c }
\hline
\multirow{2}{*}{Methods} & \multicolumn{9}{c}{Mean Chamfer Distance per point ($10^{-3}$)} \\
\cline{2-10}
{}& Avg & Airplane & Cabinet & Car & Chair & Lamp & Sofa & Table & Vessel \\
\hline\hline
3D-EPN\cite{dai2017shape} &20.147 &13.161 &21.803 &20.306 &18.813 &25.746 &21.089 &21.716 &18.543  \\
PCN-FC\cite{yuan2018pcn} &9.799 &5.698 &11.023 &8.775 &10.969 &11.131 &11.756 &9.320 &9.720 \\
PCN\cite{yuan2018pcn} &9.636 &5.502 &10.625 &8.696 &10.998 &11.339 &11.676 &8.590 &9.665 \\
TopNet~\cite{topnet2019} &9.890 &6.235 &11.628 &9.833 &11.498 &9.366 &12.347 &9.362 &8.851 \\
Ours &\textbf{8.505} &\textbf{4.794} &\textbf{9.968} &\textbf{8.311} &\textbf{9.492} &\textbf{8.940} &\textbf{10.685} &\textbf{7.805} &\textbf{8.045} \\
\hline
\end{tabular}
}
\vspace{0.7mm}
\caption{Quantitative comparison for point cloud completion on eight categories objects of ShapeNet.}
\vspace{-2mm}
\label{quantative_seen}
\end{table*}

\subsection{Point Completion on the Dataset of TopNet}\label{exp_topnet}
In this experiment, we train our model on the training data from TopNet\footnote{https://github.com/lynetcha/completion3d} and then test on our created testing data.
Since we observe that object scales of the training data are larger than scales of the testing data, we adopt random scaling augmentation technique~\cite{topnet2019} during training for all methods and the scale values are uniformly sampled between $[1/1.5,1]$.
We can see that we achieve better quantitative results for all resolutions in Table~\ref{quantitative_topnet}. 


\begin{table}[!htbp]
\centering
\small
\begin{tabular}{l|c| c |c|c}
\hline
\multirow{2}{*}{Methods} & \multicolumn{4}{c}{Resolution} \\
\cline{2-5}
{}& 2048 &4096 &8192 &16384  \\
\hline
PCN~\cite{yuan2018pcn}  &9.36 &8.17 &7.28 &6.28 \\ 
TopNet~\cite{topnet2019}  &10.23 &8.85 &7.47 &6.64 \\
Ours  &\textbf{7.61}  &\textbf{6.57}  &\textbf{5.72} &\textbf{5.21} \\
\hline
\end{tabular}
\vspace{1mm}
\caption{Quantitative comparison on the training data of TopNet. 
}
\vspace{-0.3cm}
\label{quantitative_topnet}
\end{table}

\subsection{Point Completion on Our Training Data}
In this section, we show the results on our smaller training data. Quantitative and qualitative results are shown in Table~\ref{quantitative_small} and Figure~\ref{qualitative_results_2048}, respectively. 
As shown in Table~\ref{quantitative_small}, our method outperforms both PCN and TopNet on all resolutions. The relative improvements of our method compared to PCN are $16.08\%$, $12.97\%$, $15.36\%$ and $15.56\%$ for all resolutions on our smaller training data. The improvements on our smaller training data verify the robustness and generality of our method.
We also generate 2048, 4096, 8192 and 16,384 resolution objects by training one single model on 16,384 points, and compare the results with that obtained from independent training of PCN and TopNet. We still achieve lower CD errors, which verifies the 
accuracy of our method. 
\begin{table}[!htbp]
\centering
\begin{adjustbox}{width=0.46\textwidth}
\small
\begin{tabular}{l|c| c |c|c}
\hline
\multirow{2}{*}{Resolution} & \multicolumn{4}{c}{Methods} \\
\cline{2-5}
{}& PCN~\cite{yuan2018pcn} & TopNet~\cite{topnet2019} & Ours$^*$ & Ours \\
\hline \hline
2048 & 9.02 &9.88 & 8.03 & \textbf{7.57}  \\
4096 &7.71 &8.52 &6.78 &\textbf{6.71}  \\
8192 &6.90 &7.56 &5.98 &\textbf{5.84} \\
16,384 &6.17 &6.60 &5.21 &\textbf{5.21} \\
\hline
\end{tabular}
\end{adjustbox}
\vspace{1mm}
\caption{Quantitative results on our smaller training data. We take CD ($10^{-4}$) as evaluation . Ours$^*$ represents the results obtained by using the single model trained on the 16,384 resolution output.}
\vspace{-0.3cm}
\label{quantitative_small}
\end{table}

We get three conclusions from the qualitative results in Figure~\ref{qualitative_results_2048}:
\textbf{(1)} Our method is able to generate the details not only included in the partial scan, but also for the missing parts, on both high and low resolutions. For example, the lampshade (Row 1), the empennage of the car (Row 2), and the engines of the airplane (Row 3). While both PCN and TopNet miss the detailed structure and only obtain the general object shapes.
\textbf{(2)} Our generated points are more evenly distributed. From the results of desk and chandelier (Row 4 and 5), we can see more points are located on the top surface of the desk and in the top left corner of the chandelier from PCN, while ours are evenly distributed on the object surface.
\textbf{(3)} Although we mirror the partial input with respect to the $xy$-plane, our method does not memorize the mirrored points. As the results shown in the last row of Figure~\ref{qualitative_results_2048}, our generated object is not symmetric with respect to $xy$-plane. This verifies that mirroring operation provides a initialization for the missing points, and accurate point deformations are estimated by our whole network. 
More results are shown in our supplementary material.

\subsection{Robustness to Occlusion}
To further test the robustness of the models, we manually occlude the partial inputs from testing dataset by $p$ percent of points following PCN~\cite{yuan2018pcn}, and $p$ ranges from $20\%$ to $70\%$ with a step of $10\%$. The quantitative results are shown in Table~\ref{quantitative_occlusion}. Our method achieves the best performance, although the error increases gradually as more regions are occluded. This shows that our method is more robust to noise data. More qualitative results are shown in our supplementary material. 

\subsection{Point Completion for Classification}
Following~\cite{sarmad2019rl}, we also measure the completion quality by calculating the classification accuracy on the synthesized complete point clouds. Specifically, we train one classification model by PointNet~\cite{qi2017pointnet}. The upper bound (UP) is calculated on the complete points from the testing data and the lower bound (LP) is calculated on the partial points from the testing data. The remaining values are obtained by evaluating the classification model on the generated outputs from different methods.
The quantitative results are shown in Table~\ref{cls_results}. Clearly, the complete outputs provide higher accuracy because of the defects in the partial data. Our generated results improve the accuracy by $1.59\%$ compared to PCN and TopNet, which demonstrates that our outputs are more realistic and our results preserve more accurate semantic information.
\begin{table}[!htbp]
\centering
\begin{adjustbox}{width=0.48\textwidth}
\small
\begin{tabular}{l|c| c| c|c| c| c} 
\hline
\multirow{2}{*}{Methods} & \multicolumn{6}{c}{Occlusion ratios}\\
\cline{2-7}
&20\% &30\% & 40\% & 50\% &60\% &70\% \\ 
\hline
PCN~\cite{yuan2018pcn}  &7.69 &8.84 & 10.63 & 13.30 & 17.20  &23.60 \\ 
TopNet~\cite{topnet2019}  &8.46 &9.57 &11.30 & 13.60 & 17.60 & 23.20 \\ 
Ours &\textbf{5.52} &\textbf{6.72}  &\textbf{8.46} & \textbf{11.36} & \textbf{15.26} & \textbf{21.27} \\ 
\hline
\end{tabular}
\end{adjustbox}
\vspace{0.1mm}
\caption{Quantitative comparison for occluded point clouds under different occlusion rates.  The evaluation metric is mean CD per point ($10^{-4}$).}
\vspace{-0.3cm}
\label{quantitative_occlusion}
\end{table}

\vspace{-1mm}

\begin{table}[!htbp]
\small
\centering
\resizebox{0.48\textwidth}{!}{
\begin{tabular}{l|c| c| c| c| c} 
\hline
Methods & LB & UB & PCN~\cite{yuan2018pcn} & TopNet~\cite{topnet2019} & Ours \\ 
\hline
Acc. ($\%$)  &70.50 &97.33 &92.58 &92.58 &\textbf{94.17} \\ 
\hline
\end{tabular}
}
\vspace{0.1mm}
\caption{Comparison of classification results among different methods. The upper bound (UB) represents the result tested on the complete points (ground truth) of the testing data. The lower bound (LB) represents the  result tested on the partial points of the testing data. The remaining results are obtained by the synthesized objects.}
\vspace{-0.3cm}
\label{cls_results}
\end{table}
\vspace{-1mm}

\subsection{Ablation Study}
We evaluate different components in our network, including the adversarial training, mean shape, contraction-expansion unit, mirror operation and different Chamfer Distance calculations during training. 
We denote our method without discriminator as the baseline (BS).
We use CD-P as the evaluation metric and the quantitative comparison are shown in Table~\ref{ablation}.
All experiments are done on the 2048 resolution points.
We can see that our full pipeline performs the best. Removing any component decreases the performance, which verifies that each component contributes.

\begin{table}[!htbp]
\small
\centering
\resizebox{0.48\textwidth}{!}{
\begin{tabular}{c|c| c|c|c| c} 
\hline
\multirow{2}{*}{Training loss} & \multicolumn{5}{c}{Methods}\\
\cline{2-6}
& w/o MS & w/o CE & w/o Mir & BS & w/ Dis\\ 
\hline
CD-P$^{*}$ &7.78 &7.83 &7.67 &7.67 &\textbf{7.61} \\ 
CD-P$^{\diamondsuit}$ &7.80 &7.73 &7.71 &7.68 &\textbf{7.57} \\ 
CD-T$^{*}$ &7.93 &7.90 &7.76 &7.75 &\textbf{7.68} \\ 
CD-T$^{\diamondsuit}$ &8.00 &8.01 &7.95 &7.75 &\textbf{7.62} \\ 
\hline
\end{tabular}
}
\vspace{0.15mm}
\caption{Quantitative comparisons for the ablation study. Dis represents the discriminator, MS represents the mean shape features, CE represents the contraction-expansion unit, Mir represents the mirror operation for partial points. $^{*}$ and $^{\diamondsuit}$ represent the TopNet training data and our training data, respectively.}
\label{ablation}
\end{table}
\vspace{-1mm}

\subsection{Shape Arithmetic for Feature Learning}
\begin{figure}
\centering
  \includegraphics[width=1\linewidth]{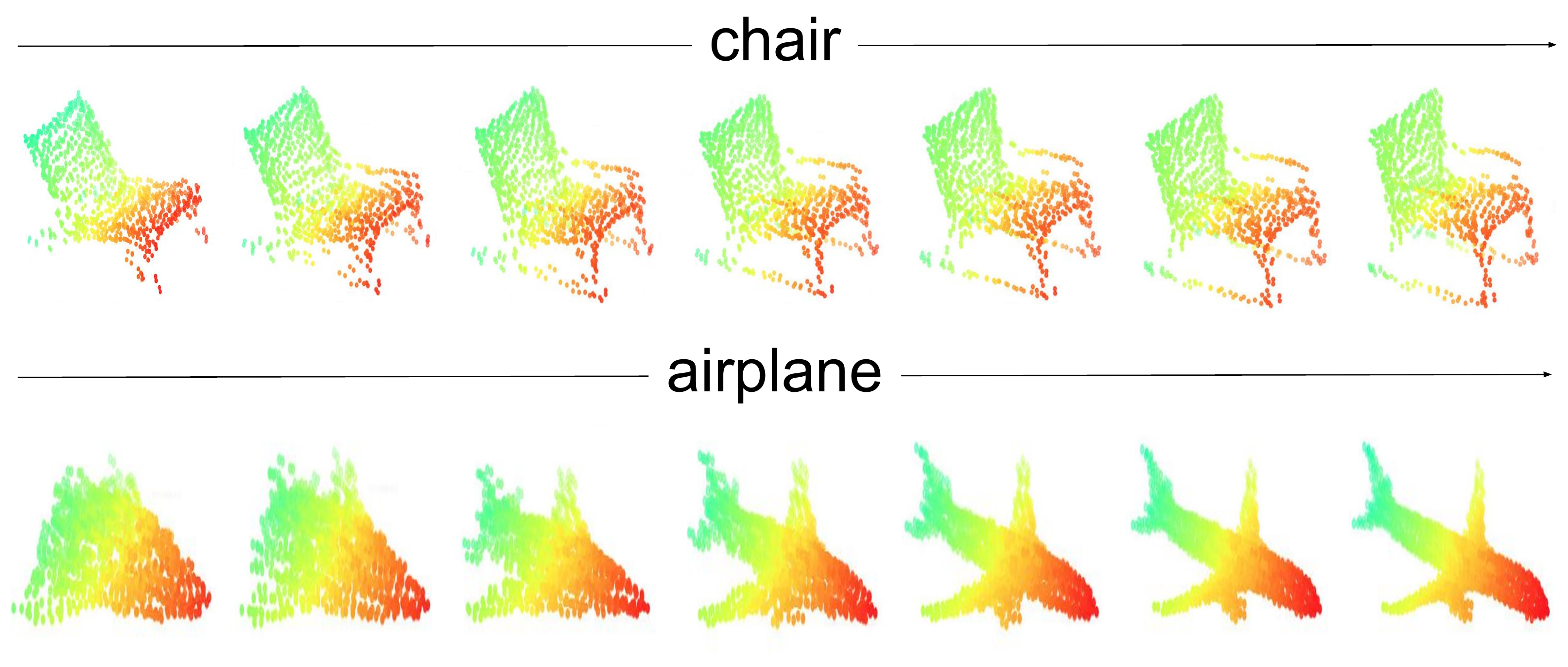}
  \caption{Shape interpolation results for chair and airplane.}
  \label{qualitative_interpolation}
  \vspace{-0.3cm}
\end{figure}

Following previous GAN methods~\cite{wu2016learning,girdhar2016learning,wang2018pix2pixHD,gulrajani2017improved,wang2017shape}, we show shape transformation by interpolating latent vectors from the encoder.
Qualitative results are shown in Figure~\ref{qualitative_interpolation}. The smooth transitions indicate that our learned features preserve critical geometric information. 
The synthesized reasonable object shapes verify the effectiveness of our cascaded refinement strategy.

\subsection{Model Size Comparison}
We evaluate the model size in Table~\ref{model_size} from two aspects for the resolution of 16,384 points: the number of parameters and the size of the trained models.
We can see that our model has fewer parameters and smaller size compared to PCN and TopNet, since we share the parameters in each cascaded refinement step.
\begin{table}[!htbp]
\vspace{-0.2cm}
\centering
\small
\begin{tabular}{l|c| c |c}
\hline
Methods & PCN~\cite{yuan2018pcn}& TopNet~\cite{topnet2019} & Ours \\
\hline
\#Paras  &6.85M & 9.96M &\textbf{5.14M}  \\
Size of Model &82.30M  &79.80M &\textbf{61.90M}  \\
\hline
\end{tabular}
\vspace{1mm}
\caption{Quantitative comparisons for model size.}
\vspace{-0.4cm}
\label{model_size}
\end{table}

\section{Conclusion}
In this work, we propose a novel point completion network to generate complete points given the partial inputs. The generator is a cascaded refinement network, which exploits the existing details of the partial input points and synthesize the missing parts with high quality. 
We design a patch discriminator that leverages on adversarial training to learn the accurate point distribution and penalize the generated objects from infidelity to the ground truth. We evaluate our proposed method on the completion datasets. Various experiments show that our method achieves state-of-the-art performances. 

\vspace{-0.3cm}
\paragraph{Acknowledgments.}
This research was supported in part by the Singapore Ministry of Education (MOE) Tier 1 grant R-252-000-A65-114, National University of Singapore Scholarship Funds and the National Research Foundation, Prime Ministers Office, Singapore, under its CREATE programme, Singapore-MIT Alliance for Research and Technology (SMART) Future Urban Mobility (FM) IRG.


{\small
\bibliographystyle{ieee_fullname}
\bibliography{egbib}
}

\end{document}